\documentclass{article}

\usepackage{microtype}
\usepackage{graphicx}
\usepackage{subfigure}
\usepackage{booktabs} 
\usepackage{hyperref}
\usepackage{amsmath}
\usepackage{amssymb}
\usepackage{mathtools}
\usepackage{amsthm}

\theoremstyle{plain}

\theoremstyle{definition}

\theoremstyle{remark}

\usepackage[capitalize,noabbrev]{cleveref}
\usepackage{subcaption}
\usepackage{algorithm}
\usepackage{algorithmic}

\usepackage[accepted]{icml2025}

\usepackage[textsize=tiny]{todonotes}

\icmltitlerunning{ADAPT: Hybrid Prompt Optimization for LLM Feature Visualization}

\begin{document}

\twocolumn[
\icmltitle{ADAPT: Hybrid Prompt Optimization for LLM Feature Visualization}

\icmlsetsymbol{equal}{*}

\begin{icmlauthorlist}
\icmlauthor{João Cardoso}{equal,ist,inesc}
\icmlauthor{Arlindo L. Oliveira}{ist,inesc}
\icmlauthor{Bruno Martins}{ist,inesc}

\end{icmlauthorlist}

\icmlaffiliation{ist}{Instituto Superior Técnico}
\icmlaffiliation{inesc}{INESC-ID}

\icmlcorrespondingauthor{João N. Cardoso}{joao.n.m.cardoso@tecnico.ulisboa.pt}

\icmlkeywords{Machine Learning, ICML}

\vskip 0.3in
]

\printAffiliationsAndNotice{}  

\begin{abstract}
Understanding what features are encoded by learned directions in LLM activation space requires identifying inputs that strongly activate them. Feature visualization, which optimizes inputs to maximally activate a target direction, offers an alternative to costly dataset search approaches, but remains underexplored for LLMs due to the discrete nature of text. Furthermore, existing prompt optimization techniques are poorly suited to this domain, which is highly prone to local minima. To overcome these limitations, we introduce ADAPT, a hybrid method combining beam search initialization with adaptive gradient-guided mutation, designed around these failure modes. We evaluate on Sparse Autoencoder latents from Gemma 2 2B, proposing metrics grounded in dataset activation statistics to enable rigorous comparison, and show that ADAPT consistently outperforms prior methods across layers and latent types. Our results establish that feature visualization for LLMs is tractable, but requires design assumptions tailored to the domain.

\end{abstract}

\section{Introduction}
\label{sec:intro}

Neural networks construct rich representations of their inputs, often assigned to linear directions in their latent activation space. Understanding the relevant input features encoded by these directions is essential for interpretability efforts, but identifying the range of input patterns that activate a given latent direction (henceforth, \textit{latent}) is a challenging task. The standard approach is to perform a dataset search \cite{paulo_automatically_2024, bills2023language}: running the model on a large corpus and registering what input examples maximally activate the directions of interest, which is an effective approach for large-scale surveys where the large computational costs of running a model on a large corpus can be amortized over thousands of latents in parallel. Dataset search is often performed for latent directions learned by sparse autoencoders (SAEs) \cite{sharkey2022taking, bricken2023monosemanticity}, which are designed to yield large dictionaries of interpretable and monosemantic latents. However, with alternate techniques for obtaining latents of interest or more complex sub-components such as model circuits or collections of latents, this approach becomes computationally prohibitive. 

An alternate approach is feature visualization \cite{olah_feature_2017}, which consists of creating an input that is ``tailor-made'' for a given latent, via prompt optimization techniques which seek to maximally activate the target latent. This brings several advantages over dataset search:
\begin{itemize}
    \item \textbf{Overcoming dataset limitations:} searched datasets are typically smaller than the model's training dataset, limiting the number of illustrative examples for any given latent.
    \item \textbf{Disentangling correlated features:} if two features happen to be correlated in the searched dataset (e.g. a semantic concept and the language of the text), it is hard to tease apart the latent's corresponding feature. \cite{olah_feature_2017}
    \item \textbf{Versatility:} any feature optimization can be modified to incorporate complex objectives and penalties.
    \item \textbf{Causal grounding:} Correlational methods can produce plausible-looking explanations even for randomly initialized networks \cite{meloux2025the}. Feature visualization constructs inputs that cause activation, sidestepping spurious associations in any fixed corpus. 
\end{itemize}

These advantages are visible in previous vision interpretability work, which has made extensive use of feature visualization. However, the application of these techniques to LLMs has been limited, mostly due to the discrete nature of textual inputs \cite{thompson_fluent_2024}, which precludes direct gradient-based updates. In this work, we take inspiration from existing adversarial prompt optimization techniques, where the objective is to induce the model to generate outputs that bypass its safety training, and from previous LLM feature visualization work. Our contributions are as follows:

\begin{itemize}
    \item We introduce ADAPT, a new algorithm that outperforms existing baselines by combining beam-search initialization with gradient-guided mutation.
    \item We evaluate ADAPT and previous methods on SAE latents from Gemma 2 2B \cite{Riviere2024Gemma2I}, introducing metrics grounded in dataset activation statistics to enable systematic comparison of different techniques.
    \item We analyze the reliability of gradient-based optimization methods for this application domain, and investigate the role of fluency penalties in prompt optimization for feature visualization.
\end{itemize}

The remainder of this paper is organized as follows. Section~\ref{sec:background} covers background on SAEs and existing prompt optimization methods. Section~\ref{sec:method} introduces ADAPT. Section~\ref{sec:experiments} presents our experimental evaluation, followed by an analysis of the common failure modes of several existing methods. In section \ref{sec:extensions} we extend ADAPT to a dual latent objective, to illustrate its versatility. We conclude in Section~\ref{sec:conclusion}.

\section{Background}
\label{sec:background}

We review sparse autoencoders, which provide our evaluation testbed, and existing prompt optimization methods, whose limitations motivate ADAPT.

\subsection{Sparse autoencoders}

The representation vectors constructed by LLMs are thought to contain far more features than they have dimensions, by encoding them as non-orthogonal directions in activation space that do not align with the neuronal basis \cite{elhage2022superposition}. Sparse autoencoders (SAEs) learn to reconstruct a target representation vector through a high-dimension middle layer \cite{bricken2023monosemanticity, templeton2024scaling}. A sparsity penalty encourages most latent dimensions to remain inactive after the activation function, yielding a small set of active directions that capture the input vector's variability. These directions have been shown to be more monosemantic, in the sense of being correlated with single input features, than individual neurons.

While the proposed ADAPT method can accommodate arbitrary directions or complex objectives, we focus on SAE latents due to the availability of open source tools that support the development of our technique, by mitigating the need to retrieve target latents ourselves, and by enabling comparison with existing latent interpretation techniques, namely dataset search, without incurring in high computational costs. In particular, the Gemma Scope project \cite{lieberum_gemma_2024} has made publicly available a suite of SAEs, with additional latent statistics and dataset examples provided via the Neuronpedia API \footnote{\url{https://neuronpedia.org/api-doc}}. 

For feature visualization of SAE latents, our objective is as follows. Given an input activation vector $x \in \mathbb{R}^n$, an SAE produces a reconstruction via:
\begin{equation}
    \hat{x} = W_{\text{dec}} \cdot f(W_{\text{enc}} x + b_{\text{enc}}) + b_{\text{dec}},
\end{equation}
where $f(x) \in \mathbb{R}^M$ is the sparse activation vector. For a target latent $i$, the optimization objective is:
\begin{equation}
    x^* = \arg\max_x f_i(x),
\end{equation}
where $f_i(x)$ denotes the activation magnitude of the $i$-th latent direction.

\subsection{Prompt Optimization}

Despite the abundance of prompt optimization methods \cite{ramnath-etal-2025-systematic}, most are not applicable to feature visualization, due to fundamental domain differences. For one, it would be infeasible to use exceedingly expensive techniques, such as those which involve LLM API calls  \cite{yang2024opro}, comprehensive tree searches \cite{wang2024promptagent}, or model fine-tuning \cite{cheng2024bpo}. A successful jailbreak prompt justifies hours of compute because it transfers across contexts, whereas feature visualization must be quick to run for individual latents since each requires individual optimization. For another, approaches that leverage prior knowledge to constrain the search space \textit{a priori} are inapplicable, given the variety of roles played by different latents. Lastly, since latents are specific to each model, surrogate models \cite{paulus2025advprompter} or simulation techniques \cite{zhang2024language} cannot be used. We thus focus on prompt methods which are self-contained, relatively inexpensive, and devoid of prior assumptions.

\subsubsection{Greedy Coordinate Gradient}

Greedy Coordinate Gradient (GCG), introduced by \citet{zou_universal_2023} optimizes directly in the discrete input space, by starting with a randomly initialized prompt and iteratively swapping out tokens. A gradient-based estimate technique, first introduced by \citet{ebrahimi-etal-2018-hotflip}, is used to constrain the token search space. The estimate is computed by determining the gradient with respect to a one-hot representation of the input prompt. This is possible due to the equivalence between the regular embedding procedure, in which an embedding matrix $W_E$ is indexed at the $i$th position, where $i$ is the token index, and a matrix multiplication $e_iW_E$, between the one-hot vector $e_i$ and the embedding matrix. Whereas indexing is non-differentiable, the multiplication procedure allows the gradient of the one-hot vector to be computed.

Starting from an initial prompt, GCG computes an estimate by converting the prompt into a $\mathrm{prompt\_length} \times \mathrm{vocab\_dim}$ matrix of one-hot vectors and running a backward pass to compute the one-hot gradients as described above. Afterwards, a random position is selected for mutation, and replacements are randomly sampled from the set of $k$ lowest gradient prompts to generate $B$ candidate prompts. The operative assumption is that the gradient value at a given index is predictive of the loss impact of discretely swapping in the associated token. Lastly, a forward pass is computed to determine the loss for each candidate, and the best performing one is retained. This procedure is repeated until a specified number iterations is reached.

\subsubsection{Beam Search}

An alternative way to get around the problems posed by input discreteness is to conduct a gradient-free tree-search over prompt space. However, given the size of typical vocabularies, the search space is prohibitively large. \cite{sadasivan_fast_2024} propose Beam Search-Based Adversarial Attack (BEAST), which explores prompt space via beam-search, where new candidates are generated by right-appending tokens sampled from the model's own predicted sequence probabilities. BEAST assumes a complex input structure and optimization objective which are not relevant for our purposes, but at its core it amounts to a simple beam-search procedure. At each step, $k_1$ prompt sequences are maintained in a beam. For each sequence, $k_2$ candidates are generated by sampling and appending one token via top-k sampling. The $k_1$ lowest loss candidates are then selected for the next beam.

BEAST's main advantage is its speed, requiring only forward passes. It thus avoids the overhead of gradient calculation. Additionally, a single forward pass serves dual purposes: evaluating candidate scores and computing logits for the next generation step.

\subsubsection{Evolutionary Prompt Optimization}

Evolutionary Prompt Optimization (EPO) is an algorithm introduced by \citet{thompson_fluent_2024} specifically for feature visualization in LLMs, which extends GCG by incorporating an evolutionary framework to balance fluency and feature activation. Rather than optimizing for a single objective, EPO maintains a population of prompts distributed along a Pareto frontier: a trade-off curve between fluency (i.e. the mean self-cross-entropy of each candidate) and feature activation strength. Each member of the population optimizes for a different weighting between these objectives, allowing EPO to discover diverse prompts ranging from fluent low-activation prompts to disfluent strongly-activating prompts.

The combined activation and fluency objective can be expressed as follows:

\[\mathcal{L}_\lambda(t) := f(t) - \frac{\lambda}{n} \sum_{i=0}^{n-1} H(\text{LLM}(t_{\leq i}), t_{i+1}) \tag{1},\]

\[t^*_\lambda = \arg\max_{t} \mathcal{L}_\lambda(t) \tag{2},\]

where $f(t)$ is the target SAE feature activation for prompt $t$; $\text{LLM}(t_{\leq i})$ denotes the language model's predicted probability distribution over next tokens given prefix $t_{\leq i}$; $H(\cdot, \cdot)$ is the cross-entropy operator; $\lambda \geq 0$ is a weighting parameter controlling the trade-off between latent maximization and fluency; $n$ is the number of tokens in prompt $t$; $t_{\leq i}$ represents the prefix of sequence $t$ up to (and including) the $i$-th token and $t_{i+1}$ is the $(i+1)$-th token in $t$.

The second term computes the average self-cross-entropy of the sequence, which 
is equivalent to the negative log-likelihood under the language model. Lower 
cross-entropy corresponds to more `fluent' (i.e. higher likelihood) text.

A few expansions of EPO have been proposed. \citet{thompson_flrt_2024} extend EPO with 3 additional operations: insertions, logit-swaps, deletions; where the former two sample new tokens from the model's own predicted sequence probabilities. While these extensions were made for an adversarial context, they are easily transferable to the feature visualization domain.

\citet{shabalin_evolutionary_2024} apply EPO to SAE features obtained from the Gemma 2 2B model, and show that it carries over well, producing moderately interpretable prompts that activate the target latents.

\citet{graham2025context} present ContextBench, a benchmark for fluent context modification targeting latent activation and behavior elicitation, and introduce two extensions to EPO. EPO-Assist periodically queries a large language model to generate novel variations of current outputs, which are then incorporated back into the optimization population. EPO-Inpainting identifies tokens with maximum per-token activation, freezes them, and uses a bidirectional diffusion model (LLaDA) to regenerate surrounding tokens, preserving optimization progress while enhancing fluency. 

The authors curate 102 features from Gemma 2 2B across three taxonomic axes: activation density (how frequently the feature fires), vocabulary diversity (semantic breadth of activating concepts), and locality (whether activation is concentrated on single tokens or distributed across sequences), and use it to benchmark and validate the developed methods. 

\section{ADAPT: Adaptive DynAmic Prompt Tuning}
\label{sec:method}

We describe ADAPT's three main components: beam-search initialization (\ref{sec:initialization}), candidate generation via hybrid mutation (\ref{sec:candidate_generation}), and diversity-preserving evaluation and selection (\ref{sec:evaluation_culling}).

\subsection{Overview}

Our proposed method is an extension of GCG which incorporates several elements that overcome limitations we observed when applied to feature visualization. In this section, we present the method's core architecture, leaving validation of key design decisions to section \ref{sec:validation}. An illustrative diagram for ADAPT is shown in Figure \ref{fig:adapt_diagram}, and code is included in the supplementary materials.

\begin{figure} 
    \centering 
    \includegraphics[width=1\linewidth]{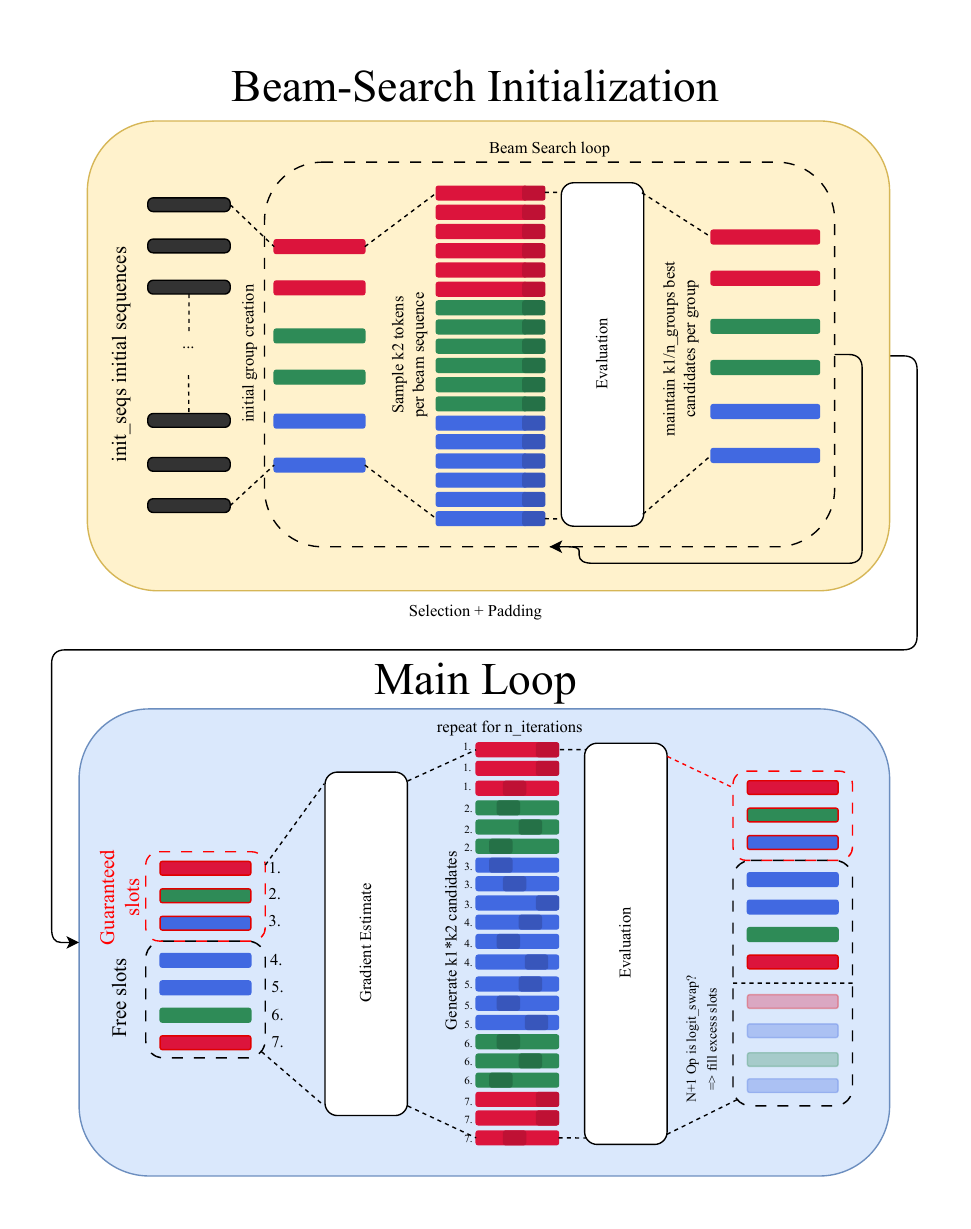} 
    \caption{Diagram for ADAPT.} 
    \label{fig:adapt_diagram}     
\end{figure}

\subsection{Initialization}
\label{sec:initialization}

During experiments with GCG, we found that the choice of initialization strongly affected whether a highly activating prompt was successfully found for a given latent. We also noticed that the optimization trajectory resulting from a given initialization was crucial, with prompts optimized in parallel often achieving drastically different final losses (see subsection \ref{sec:validation}).

As such, ADAPT initializes via a beam-search procedure, where several beams are maintained and optimized in parallel, foregoing the need to initialize with an arbitrary user-defined or randomly generated input sequence. Starting from a number of single-token sequences, candidates are generated by right-appending likely next tokens, after which point the lowest loss candidates are included in the next iteration's beam. The number of beams \texttt{n\_groups}, the number of initial sequences, the beam-size $k1$, and the expansion factor $k2$ are specified by the user. 

We observed that the structure of SAE-latent maximization does not lend itself well to sole use of right-append operations, since we take the latent activation magnitude at the last position, which is continuously relocated to that of the right-appended token. As a hypothetical example, for a latent that exclusively fires for the token \texttt{foo}, a right-append-only beam search would converge on the prompt ``\texttt{$\cdots$ foo foo foo}'', whereas a better prompt might be obtained by refining the left-hand context instead. As such, the initialization procedure also performs middle-inserts, by randomly selecting a sequence position and sampling from the model's own predicted distribution at that position.

Each beam is independent, which means that our initialization is equivalent to a parallel run of more than one beam search procedure, each with its own initial starting set. The aim is to enable better exploration of the search space, decreasing the likelihood of stagnating in suboptimal local minima. The initialization returns the best prompts across all beams and iterations, without losing track of beam provenance, so that no cross-contamination takes place. Since each iteration increases the sequence length, but the initialization returns the best prompts achieved across all iterations, we may start the main optimization loop with prompts of differing lengths, requiring appropriate masking.

\subsection{Candidate Generation}
\label{sec:candidate_generation}

ADAPT optimizes over an adaptively sized population of prompts, initialized to the best sequences returned by the beam-search initialization. At each iteration and for each prompt, a fixed, user-specified number of candidates is generated via one of two mutation mechanisms: GCG-style swaps, using the aforementioned gradient-based estimate, and logit-swaps, as in EPO. We multinomially sample the next iteration's operation based on a user-specified probability value, and retain an excess number of candidates in the surviving population if the next operation is a logit-swap, in order to take advantage of the lower computational requirements of the gradient-free logit-swap mutation. Mutation positions are selected either randomly or by sampling from a distribution over sequence positions that biases towards the selection of later positions. Since the initialization may return underlength prompts, we instead perform middle-inserts for logit-swap iterations. 

\subsection{Evaluation and Culling}
\label{sec:evaluation_culling}

ADAPT maintains a complex slot management system. First, each prompt has a group ID, carried over from the initialization procedure. No cross-group selection takes place, to avoid rapid convergence on minor variations of the same prompt, thereby aiding in exploration. Not only does this serve to minimize the likelihood of getting prematurely stuck in local minima, but it also provides diverse, independent examples for a single latent. Notwithstanding, the user may define a \texttt{mergepoint} after which the selection mechanism defaults to a greedy global selection that disregards group provenance. Second, two types of slots exist. The first are group-specific ``guaranteed'' slots, which always contain the group's best scoring prompt across all iterations. The second set comprises ``free'' slots, which are allocated via global selection of the highest-improvement candidates, where the improvement is measured as the difference between the achieved loss and the previous group best. This dual structure maintains diversity through the guaranteed slots, while allowing high-performing mutations from any group to claim additional representation through the free slots. With that said, we cap each group at twice its proportional share of free slots.

Before the candidate selection phase can occur, each one must be evaluated. Our evaluator begins by running a forward pass on the candidates, storing both the latent activations and the logits. Afterwards, we apply a fluency penalty mechanism. We employ a global sigmoid fluency penalty schedule, which gradually ramps up depending on several parameters. The penalty weight at any given time $\lambda_t$ is a function of the maximum penalty weight $\lambda_{\max}$ that was applied, the iteration of steepest increase $t_{\text{mid}}$, and the parameter $s$, which controls the steepness of the increase:

\[ \lambda_t = \frac{\lambda_{\text{max}}}{1 + \exp(-s \cdot (t - t_{\text{mid}}))}.\]

The user may also choose a fluency threshold. Prompts with a mean cross-entropy score below this threshold do not incur any fluency penalty, while prompts above this threshold get penalized in proportion to how much their mean self-cross-entropy exceeds it. We allow the user to fully customize the fluency penalty strategy by setting the cross-entropy threshold to 0 (disabling penalty thresholding), or the steepness $s$ to 0  (disabling the fluency schedule entirely). We analyze the importance of fluency in Section \ref{sec:fluency}.

\section{Experiments}
\label{sec:experiments}

Feature visualization for SAE evinces the strengths and weaknesses of existing techniques, which informed our design decisions in ADAPT. With the experiments presented in this section our objectives were twofold:
\begin{enumerate}
    \item Comparatively evaluating 4 different methods---ADAPT, GCG, BEAST, and EPO---in order to establish the superiority of our method as a whole.
    \item Validating our design decisions through tailored individual experiments. The results illustrate the failure modes of different techniques, putting into question assumptions held in previous work.
\end{enumerate}

\subsection{Setup}
We first describe implementation details and hyperparameter settings for all evaluated methods (\ref{sec:implementation}), then our model and feature selection procedure (\ref{sec:model_selection}), and finally the evaluation metrics used for comparison (\ref{sec:evaluation_metrics}).

\subsubsection{Implementation details}
\label{sec:implementation}

Hyperparameters were set to sensible defaults according to the authors' own implementations. Individual optimization runs ran in approximately 80s in optimal GPU conditions, for a single NVIDIA A40 48GB GPU setup. 

Our implementation of GCG optimizes 6 prompts in parallel, with 32 candidates generated at each iteration. The highest achieved activation is reported, unless stated otherwise. 

BEAST was run with a beam size of 128 and expansion factor of 32. The beam size is far higher that GCG's candidate count due to the lower iteration count (10, corresponding to the prompt length), and the absence of backward passes.

We took Thompson et al.'s own implementation as the foundation of our implementation of EPO. Identical hyperparameters were set to match GCG's as far as possible, with a few key differences. The fluency penalty was set to range from 0.1 to 15 across the 6 slots. The prompt lengths were bounded between 8 and 15, because of the algorithm's use of length-altering mutations. Operation probabilities were set as follows: GCG-swap (0.5), logit-swap (0.2), insert (0.15), and delete (0.15). 

ADAPT was configured with 10 prompts maintained across 3 diversity groups, with 32 candidates generated per iteration. For the beam-based initialization phase a beam size of 30 and an expansion factor of 32 were used. During optimization, GCG-swap and logit-swap operations were applied with equal probability (0.5 each). Each group retained two guaranteed slots; the remaining 4 or 14 slots (depending on iteration type) were allocated competitively. Cross-entropy regularization was applied with a maximum penalty of 3.0. The GCG estimator sampled from the top 512 tokens ranked by gradient magnitude. Target prompt length was fixed at 10 tokens, and optimization ran for 25 iterations.

We sought to maintain comparable prompt lengths across all methods, despite EPO's variable length implementation. Further details are available in Appendix \ref{sec:appendix_implementation}.

\subsubsection{Model and Feature Selection}
\label{sec:model_selection}

Our experiments were performed on Gemma 2 2B \cite{Riviere2024Gemma2I}, in keeping with previous work \cite{shabalin_evolutionary_2024}. We used the 16k-wide residual stream SAE suite from Gemma Scope \cite{lieberum_gemma_2024}. The residual stream represents the cumulative computation accessible to all downstream layers, making it the natural target for understanding what information the model maintains and routes through its depth.

A cursory glance at any set of dataset examples for different latents will reveal that they vary widely in their activation patterns and roles. While some activate for specific tokens, others respond to a wide range of tokens that share some functional or semantic role. Some latents are active for longer subsequences, or activate more frequently than others. Accordingly, we designed our latent selection to capture variance along these axes. 

We curated three latent sets of differing sizes. The larger latent set contains 486 latents sampled from 6 evenly spaced layers (layers 3, 7, 11, 17, 21, and 25) across the model's 26-layer depth. We began by classifying the first 1000 latents for each layer according to the the items in the taxonomy described by Graham et al.---locality (the narrowness of the range of tokens the latent is active for in the context), diversity (how many different tokens achieve maximal activation across several examples), and density (how frequently the latent activates in the whole dataset). We implemented computational proxies for these items by leveraging information available via the Neuronpedia API; our procedure is detailed in Appendix \ref{sec:appendix_latent_selection}. Subsequently, we performed stratified sampling, considering the first 1000 latent directions for each SAE, by applying a $3 \times 3 \times 3$ percentile grid based on the density, locality, and diversity scores. We sampled 4 features from each bin, yielding a total of 81 per layer.

\subsubsection{Evaluation}
\label{sec:evaluation_metrics}

\begin{figure}[!htb]
    \centering
    \includegraphics[width=1\linewidth]{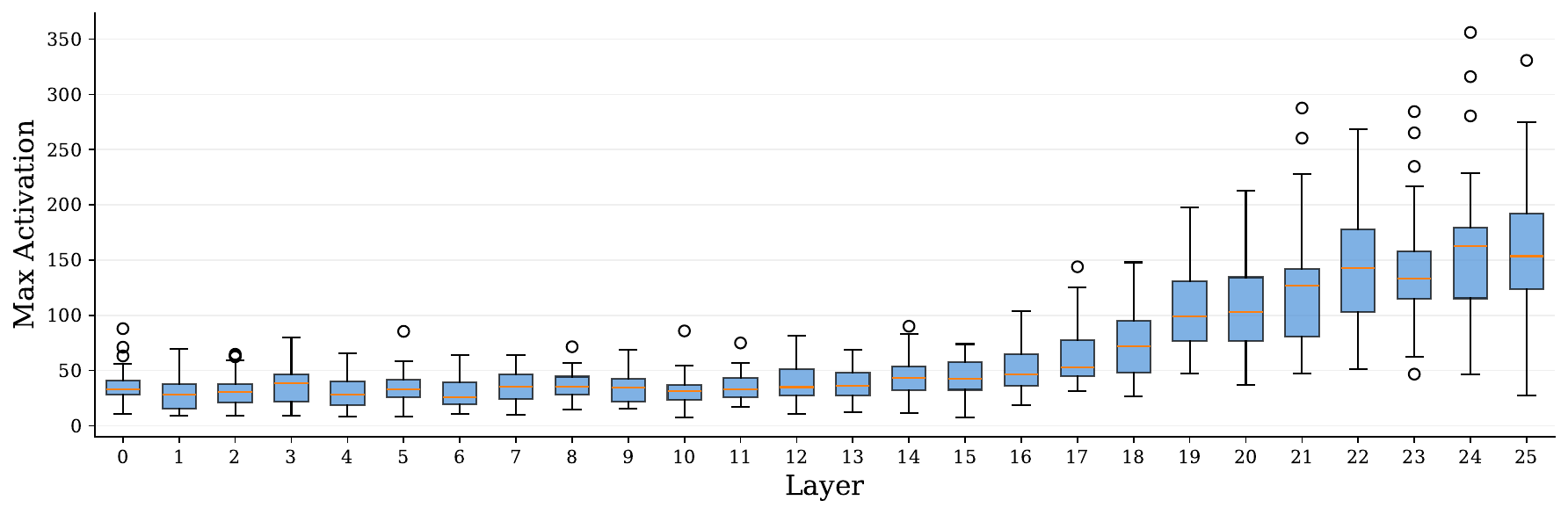}
    \caption{Distribution of top dataset activations for latents 0–49 (by Neuronpedia index) across all layers of Gemma 2 2B.}
    \label{fig:act_dist}
\end{figure}

Despite EPO's and ADAPT's use of fluency penalties, we consider the raw activation values in our metrics, relegating our discussion of fluency to section \ref{sec:validation}. Since activation magnitudes differ from latent to latent, and also tend to vary across layers---typically, later layers contain latents with much higher Maximally Activating Examples (MAEs), as seen in Figure \ref{fig:act_dist}---our metrics compare the optimized prompt's activation with the corresponding MAEs in two ways. The first metric is the \textit{activation ratio}: the ratio between the optimized prompt and the top MAE. The second metric is the \textit{activation rank}, which corresponds to the position the optimized prompt would occupy if ranked among the top 15 de-duplicated MAEs. A prompt with a rank of 1 outperforms all MAEs and, consequently, its activation ratio is greater than 1, while a prompt with rank 16 is worse than all 15 MAEs.   

\subsection{Overall Performance}
\label{sec:performance}

\begin{figure}[!htb]
    \centering
    \includegraphics[width=0.7\linewidth]{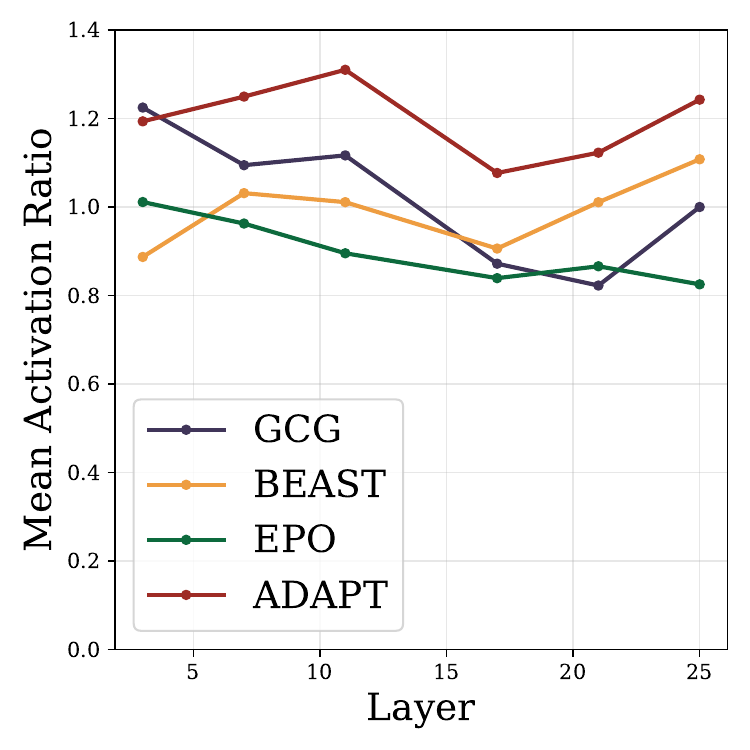}
    \caption{Mean activation ratio (the optimized prompt's activation divided by the maximally activating dataset example) per layer, for all 4 methods under study.}
    \label{fig:mean_ratio}
\end{figure}

\begin{table}[h]
\centering
\begin{tabular}{c|cccc}
 & GCG & BEAST & EPO & ADAPT \\
\hline
GCG   & --   & 56.4\% & 63.2\% & 29.6\% \\
BEAST & 43.6\% & --   & 50.2\% & 12.8\% \\
EPO   & 36.8\% & 49.8\% & --   & 21.4\% \\
ADAPT & 70.4\% & 84.0\% & 78.6\% & --   \\
\end{tabular}
\caption{Row beats column comparison for all 4 methods, considering all 486 latents.}
\label{tab:method_comp}
\end{table}

Fig. \ref{fig:mean_ratio} shows the mean activation ratio per considered layer and for all methods, showing that ADAPT outperforms or is on par with every other method on all layers.  Among the previous methods, GCG is dominant for the first layers, while BEAST edges it out on the last layers, indicating a degrading performance profile for GCG toward later layers. Furthermore, the failure modes of GCG and BEAST appear to be complementary, which is corroborated by the fact that ADAPT, which incorporates both a beam-search procedure and gradient-based mutations, achieves a more constant performance profile throughout the layers.

The comparison in Table \ref{tab:method_comp} shows that ADAPT achieves higher activation than GCG on 70.4\% of features (Wilcoxon signed-rank $p < 10^{-28}$), further corroborating the improved performance profile.

EPO underperforms throughout, despite the complexity of its design. This is to be expected to a certain extent, since EPO employs a global selection mechanism which rapidly converges on minor variations of the same prompt spanning the pareto frontier between activation and fluency, whereas our implementation of GCG devotes the computational budget to optimizing 6 prompts independently. In other words, EPO can be seen as an analogue to a single-slot GCG optimization procedure which devotes additional resources to exploring fluent variations which may not vary significantly among themselves, which we are then comparing to the maximum GCG activation across 6 independently-optimized prompts. While this does not undermine the overall approach, it is underpinned by an assumption about the importance of fluency, which we separately revisit in a later section. Since it is the only design element EPO exclusively contains, we forego further analysis of EPO in this section.

\begin{figure}[t]
  \centering
  \begin{minipage}{0.49\columnwidth}
    \centering
    \includegraphics[width=\linewidth]{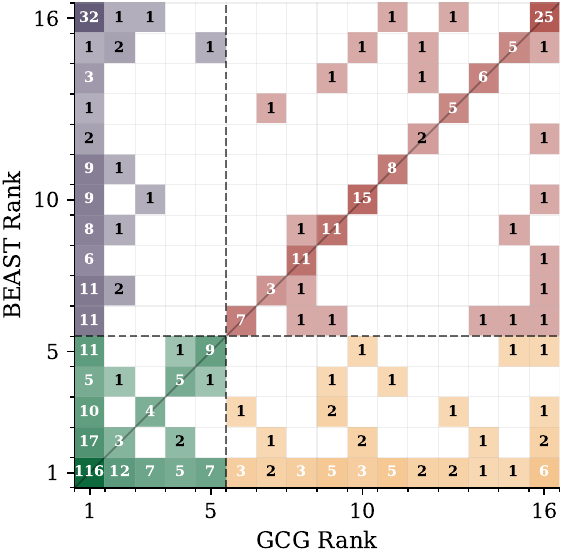}

    \small GCG vs BEAST
  \end{minipage}\hfill
  \begin{minipage}{0.49\columnwidth}
    \centering
    \includegraphics[width=\linewidth]{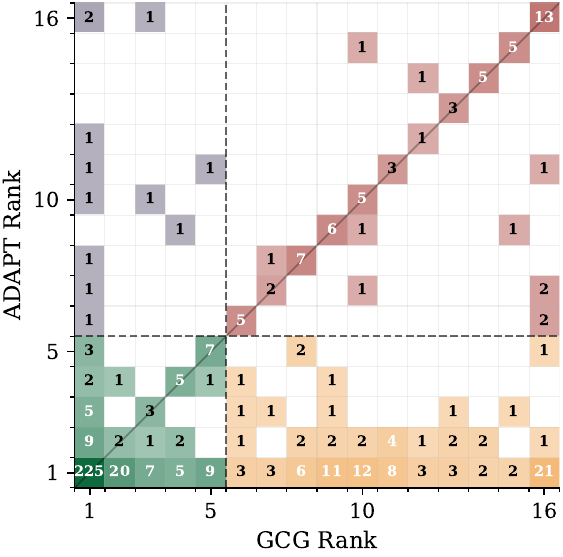}

    \small GCG vs ADAPT
  \end{minipage}
  \caption{Comparative ranking matrices
for GCG-ADAPT and GCG-BEAST comparisons.}
  \label{fig:rank_matrices}
\end{figure}

\begin{figure}[t]
  \centering
  \begin{minipage}{0.49\columnwidth}
    \centering
    \includegraphics[width=\linewidth]{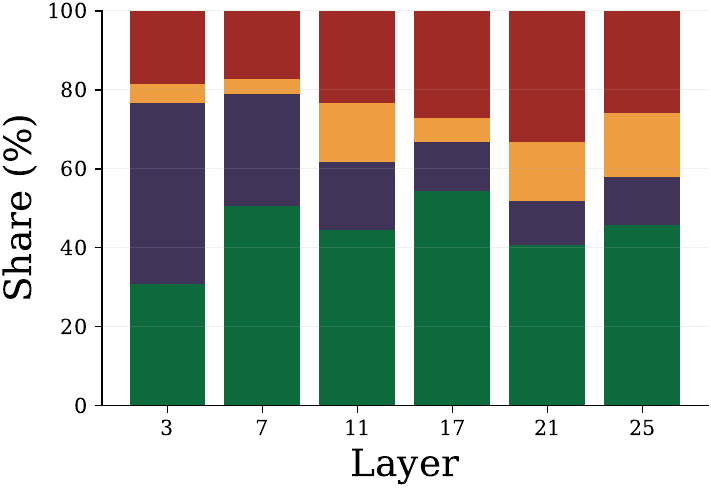}

    \small GCG vs BEAST
  \end{minipage}\hfill
  \begin{minipage}{0.49\columnwidth}
    \centering
    \includegraphics[width=\linewidth]{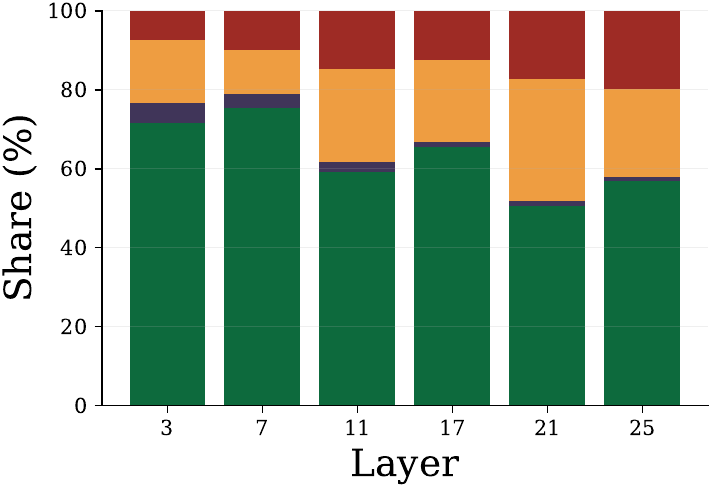}

    \small GCG vs ADAPT
  \end{minipage}
  \caption{Quadrant success per layer for GCG-BEAST and GCG-ADAPT comparisons.}
  \label{fig:quadrant_shares}
\end{figure}

Figure \ref{fig:rank_matrices} shows two activation rank confusion matrices for the GCG-BEAST and GCG-ADAPT comparisons. Arbitrarily considering that a rank greater than 5 corresponds to a successful optimization, the green and red quadrants indicate joint success and failure, respectively, while the other quadrants indicate asymmetric success. The stacked bar charts in Figure \ref{fig:quadrant_shares} indicate the per-layer share of successful vs. unsuccessful features. The GCG-BEAST comparison further indicates that their failure modes are complementary, with a significant share of GCG-only successes toward earlier layers. The GCG-ADAPT comparison shows that the share of GCG-only successes is marginal, while ADAPT-only successes are substantial, further establishing the superiority of the method. 

\subsection{Validation Experiments}
\label{sec:validation}

The overall performance results indicate that the domain is challenging for existing methods. While the target loss function is continuous, a successful optimization run should ideally find the pattern encoded by the latent, which, given the SAE reconstruction objective, is expected to be discrete and monosemantic. For particularly local and low-density features, this pattern might consist of a single vocabulary token, lending a "needle-in-a-haystack" quality to the task. Additionally, latents are, by definition, non-orthogonal, which means that there can be interference between them, especially for prompts eliciting lower activation magnitudes. In this section, we analyze the effectiveness of existing techniques at finding the key feature patterns, as well as the relationship between our metrics and human interpretability.

In order to do so, we curate two smaller latent sets, each with 50 latents sampled from the larger dataset, so as to facilitate targeted experiments. The first is a ``balanced'' latent set, where we sample features from every layer, seeking to maintain a representative difficulty profile. The second is a ``hardness-weighted'' latent set, for which we sample 50 features from the last 3 layers, which contain harder latents to optimize (for GCG, specifically), and skew our selection towards more difficult latents.

In order to assess hardness, we determine the inverse of the average activation ratio obtained by GCG for every iteration, averaged across all 6 prompts. The average activation ratio is equivalent to the ratio between the area under the trajectory curve and the area below $y=a_{\max}$, $a_{\max}$ being the maximal dataset activation. Intuitively, an easy latent is one for which GCG rapidly converges on a solution that is equal to or greater than $a_{\max}$ across all 6 slots, in which case the AUC ratio will be 1 or lower. If all trajectories plateau below the dataset reference maximum, then the hardness will be far greater than 1. The hardness $H_l$ for a latent $l$ is given by:

$$
H_l = \left(\frac{1}{S \cdot a_{\max} \cdot T} \sum_{s=1}^{S} \sum_{t=1}^{T} \max(L_s^{(t)}, 0)\right)^{-1},
$$

where $S$ is the number of prompt slots, $T$ is the number of iterations, and $L_s^{(t)}$ is the loss (negative activation) for slot $s$ at iteration $t$.

We divided latents into 5 hardness quintiles based on percentile boundaries, and sampled features from each quintile according to two strategies. The uniform sample drew 10 features from each quintile (50 total), while the hardness-weighted sample emphasized harder features with counts of (5, 8, 10, 12, 15) from the easiest to the hardest quintiles, respectively (50 total).

\subsubsection{The GCG estimate's signal-to-noise ratio}
\label{sec:gcg_snr}

The gradient-based estimate used in GCG (henceforth, the `GCG estimate') is predicated on the assumption that the loss impact of discretely including each token is well-approximated by its corresponding gradient computed for the one-hot encoding. If the GCG estimate were a perfect approximation, we would expect low-ranked tokens---the tokens with the most negative gradient---to lead to the greatest loss decreases. Conversely, we would also expect the tokens resulting in the largest loss decreases to have the lowest gradients.

In order to assess these two claims, we perform the following experiment on the balanced latent set. For each latent, we begin by determining the two slots which achieved the highest and lowest final activation, respectively. For each slot, we obtain the cached prompt trajectory from the main run. Afterwards, we generate 128 new candidate mutations in two ways. The first 64 are generated via the same GCG estimate procedure as in the original GCG, with a top-$k$ of 1024 (doubling the candidate count and top-$k$). The last 64 are generated by instead performing top-$p$ sampling from the model's own probabilities, with $p=0.95$. We keep track of all the generated candidates which constitute improvements over the cached prompt's activation. We also fix the trajectory to ensure comparable results between the two candidate generation methods.

\begin{figure}[ht]
    \centering
    \begin{minipage}{0.48\linewidth}
        \centering
        \includegraphics[width=\linewidth]{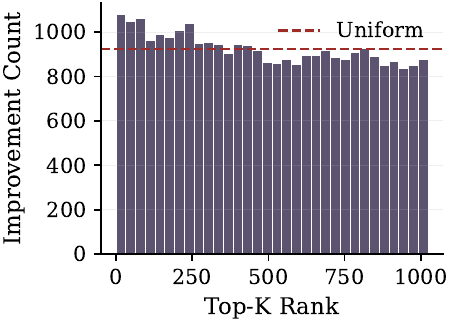}
        \caption{Number of improvements per estimate rank bin.}
        \label{fig:gcg_topk}
    \end{minipage}\hfill
    \begin{minipage}{0.48\linewidth}
        \centering
        \includegraphics[width=\linewidth]{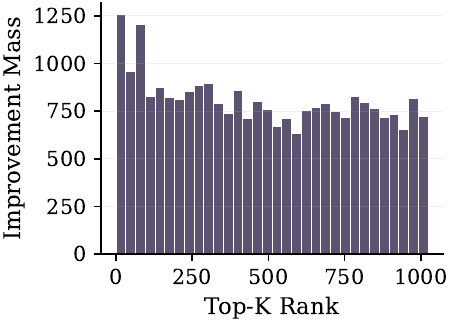}
        \caption{Improvement mass per estimate rank bin.}
        \label{fig:gcg_topk_mass}
    \end{minipage}
\end{figure}

Figure \ref{fig:gcg_topk} shows that the gradient rank distribution for improvements obtained via GCG is quite close to uniform, indicating that lower gradient values do not yield remarkably more improvements. Figure \ref{fig:gcg_topk_mass}, on the other hand, shows that the distribution of the sum of improvements is more uneven, with the lower gradient bins containing a disproportionate share of the absolute improvements. This suggests that, while the GCG estimate is noisy, and a large share of the sampled candidates are only marginally better than would otherwise be expected, it is able to occasionally yield disproportionately good candidates among its highest ranks.

\begin{figure}[ht]
    \centering
    \begin{minipage}{0.48\linewidth}
        \centering
        \includegraphics[width=\linewidth]{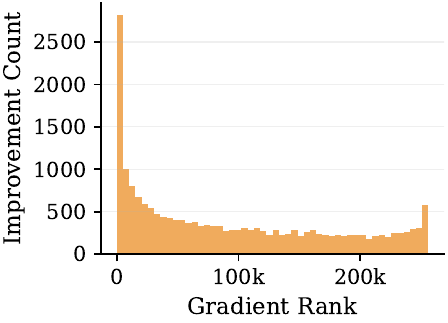}
        \caption{GCG estimate rank distribution for improvement candidates obtained via the logit-swap procedure.}
        \label{fig:logit_ranks}
    \end{minipage}\hfill
    \begin{minipage}{0.48\linewidth}
        \centering
        \includegraphics[width=\linewidth]{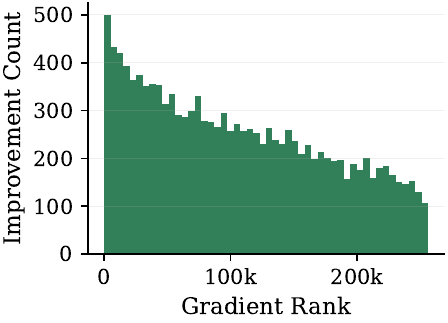}
        \caption{GCG-Estimate rank distribution for improvement candidates obtained via the random sampling procedure.}
        \label{fig:random_ranks}
    \end{minipage}
\end{figure}

Figure \ref{fig:logit_ranks} shows the gradient rank distribution for candidates selected via gradient-free logit sampling. While the lowest gradient bins capture a disproportionate share of the improvements, the distribution is close to uniform for the middle bins, indicating that there are many improvements which the GCG estimate fails to attribute an appropriate gradient to. Figure \ref{fig:random_ranks} further illustrates this, showing the distribution of the gradient ranks of tokens sampled uniformly from the entire vocabulary, with lower-gradient ranges containing more improvements. The bimodal distribution for the logit-swap candidates nonetheless suggests an interesting relationship between predicted probabilities and the biases of the GCG estimate. Specifically, the GCG estimate appears to be biased towards certain commonly occurring tokens more than would be expected by chance, while systematically undervaluing certain tokens, even when they bring about loss decreases. Inspecting the sampled tokens from the last bins, we found that they mostly correspond to punctuation and formatting tokens. 

\subsubsection{Consistency of GCG}

While the experiments in Subsection \ref{sec:gcg_snr} establish the poor signal-to-noise ratio of GCG, with many good candidates being underestimated, and many highly estimated candidates underperforming, questions remain as to how reliably GCG can converge on high activation regions of prompt space. For this reason, we now analyze how sensitive GCG is to initialization and trajectory.

For the benchmark in Section \ref{sec:performance}, GCG was configured to optimize six prompts in parallel. We observed a good deal of inter-slot variability in final activation. In this subsection, we consider the impact of initialization on the performance of GCG, by answering two questions:

\begin{itemize}
    \item Do some specific initial prompts reliably predict higher final activations?
    \item Does GCG consistently attain similar activations across different runs initialized with the same prompt? 
\end{itemize}

To answer these questions, we conduct a simple experiment. We begin by determining which of the 6 initial prompts in the main benchmark achieved the both the highest and lowest average final activations. Since the highest-performing prompt was obtained through autoregressive generation, and the lowest-performing prompt consists of randomly sampled tokens from the vocabulary, we call them the ``autoregressive'' and ``random'' prompts, respectively. Afterwards, we conduct a benchmark on the hardness-weighted latent set, where 3 of the 6 initial slots are initialized with the autoregressive prompt, and the remaining 3 are initialized with the random prompt.

\begin{figure}[ht]
    \centering
    \begin{minipage}{0.48\linewidth}
        \centering
        \includegraphics[width=\linewidth]{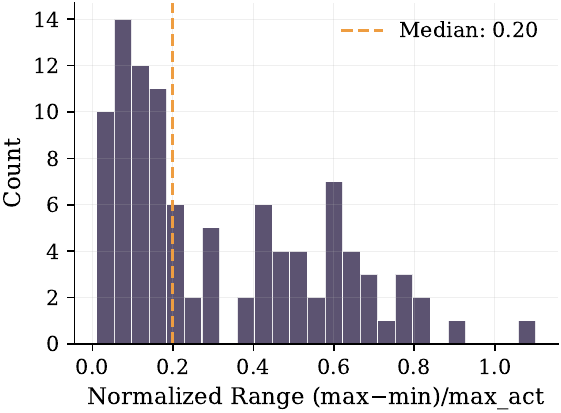}
        \caption{Distribution of activation variability across runs from identical initializations, normalized by the MAE.}
        \label{fig:intra_prompt}
    \end{minipage}\hfill
    \begin{minipage}{0.48\linewidth}
        \centering
        \includegraphics[width=\linewidth]{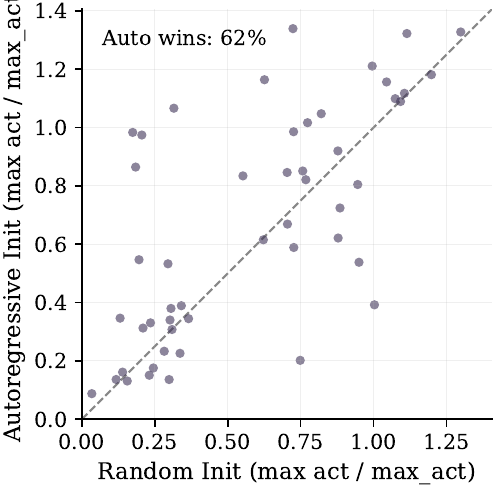}
        \caption{Comparison of best activations achieved from autoregressive versus random initializations. Points above diagonal indicate autoregressive advantage.}
        \label{fig:inter_prompt}
    \end{minipage}
\end{figure}

Figure \ref{fig:intra_prompt} shows the intra-prompt distribution of final activation ranges, normalized by the maximal dataset activation. It is clear that for both prompts there is significant variability, meaning that the same initialization can yield very different results depending on the trajectory, which is selected stochastically. This indicates that GCG is prone to falling in local minima depending on the specific trajectory through prompt space.

Figure \ref{fig:inter_prompt} shows the comparison between the performance for both slots across all slots, with the mean and max normalized activations, respectively, in each plot. It illustrates the inter-prompt variability by establishing that the autoregressive prompt slots consistently obtain higher activations than the random prompt.

Taken together, these results establish that GCG's performance is heavily dependent on location in prompt space. In other words, for some starting locations, GCG is significantly more likely to converge on high-activating token patterns, although the greedy candidate selection often produces trajectories that evolve towards local minima, highlighting the importance of open-ended exploration.

Figure \ref{fig:intra_prompt} shows the distribution of activation ranges across runs from identical initializations, normalized by maximal dataset activation. The median variability of 0.20 indicates that runs from the same starting point can diverge substantially, suggesting GCG is prone to converging on different local minima depending on the stochastically-selected trajectory through prompt space.

Figure \ref{fig:inter_prompt} compares best activations achieved from autoregressive versus random initializations. Autoregressive initialization outperforms random initialization in 62\% of configurations, demonstrating that the starting location in prompt space meaningfully influences optimization outcomes, but does not fully predict them.

Taken together, these results establish that GCG's performance depends heavily on both initialization and trajectory. While autoregressive initialization provides a consistent advantage, the substantial intra-prompt variability highlights the importance of maintaining diverse exploration to avoid local minima.
 
\subsubsection{The Role of Fluency}
\label{sec:fluency}

GCG usually produces `gibberish' prompts, which contain incoherent token sequences mixing languages, scripts, punctuation, and subword fragments. This is due to the lack of any constraint on possible token sequences. In the adversarial domain, fluency is usually enforced to bypass perplexity-based filters and to obtain in-distribution prompts. However, it is not immediately apparent that the same desiderata apply to feature visualization. For one, perplexity filters have no analogue in this domain; for another, adversarial behavioral objectives admit multiple sufficient mechanisms (e.g. simultaneously suppressing refusal directions and hijacking attention \cite{arditi2024refusal}), whereas targeted latent activation objectives specify the intervention site directly, achieving mechanistic specificity by construction rather than requiring auxiliary constraints.

Given the attributes of the feature visualization domain, two putative roles for fluency are relevant:
\begin{itemize}
    \item \textbf{Regularization:} preventing optimization from getting stuck in local minima, resulting in the return of low activation prompts.
    \item \textbf{Interpretability:} preventing high activation solutions that are uninterpretable or unrelated to the feature pattern \cite{thompson_fluent_2024}.
\end{itemize}
\begin{figure}
    \centering
    \includegraphics[width=0.8\linewidth]{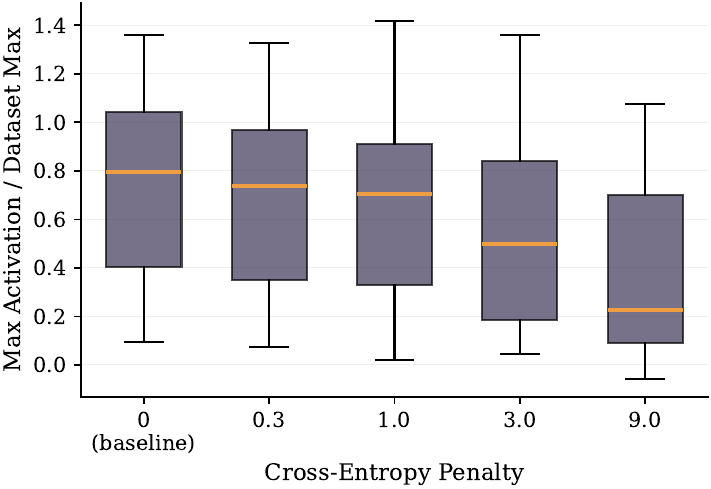}
    \caption{Activation ratio distribution for GCG runs with varying fluency penalty weights.}
    \label{fig:gcg_fluency_runs}
\end{figure}
The first hypothesis suggests that a fluency penalty may reshape the loss landscape, smoothing over narrow valleys and sharp local minima that trap optimization in gibberish regions. To test this, we rerun GCG on the hardness-weighted latent set with an added self-cross-entropy penalty:
$$
\mathcal{L}(t) = -f(t) + \frac{\lambda}{n} \sum_{i=0}^{n-1} H(\text{LLM}(t_{\leq i}), t_{i+1}),
$$
where $f(t)$ is the target latent activation, $H(\cdot, \cdot)$ is the cross-entropy, and $\lambda \in \{0.3, 1.0, 3.0, 9.0\}$ controls the fluency-activation tradeoff. 

Figure \ref{fig:gcg_fluency_runs} shows that higher values of $\lambda$ hinder optimization across the whole latent set, thereby undermining the hypothesis. Notwithstanding, we highlight that this experiment does not directly bear on the fluency configuration used in the ADAPT benchmark, since higher penalties are only applied in later iterations, leaving the initialization and early optimization periods relatively unhindered by fluency.

To assess whether fluency penalties improve interpretability and to validate ADAPT's design, we conducted a manual evaluation comparing three conditions: GCG, GCG with a fluency penalty ($\lambda=1.0$), and ADAPT. Three raters (including one author) evaluated prompts on 60 latents: the 50-latent hardness-weighted set plus 10 additional latents with activation ratio greater than 1.0, included to better probe whether high-activation prompts recover interpretable patterns. For each latent, raters viewed the latent's explanation, maximally activating dataset examples, and three prompts (one per condition, order randomized). Since the patterns captured by certain latents can be hard to infer, consisting of upweighted predictions which are not necessarily borne out in the dataset examples, structural properties of the text, or otherwise unintuitive patterns, the explanations were generated by the authors, although a set of maximally activating examples was provided\footnote{We have included the annotation file in the supplementary materials.}. Prompts were rated on a 1–5 scale: (1) no discernible pattern, (2) semantic ballpark but not discernible, (3) pattern understandable but not evident from prompt alone, (4) pattern present but differs from examples (e.g., different language), (5) pattern obvious and matches examples. We computed Krippendorff's $\alpha$ for inter-rater agreement, obtaining $\alpha=0.650$.
\begin{figure}
    \centering
    \includegraphics[width=1.0\linewidth]{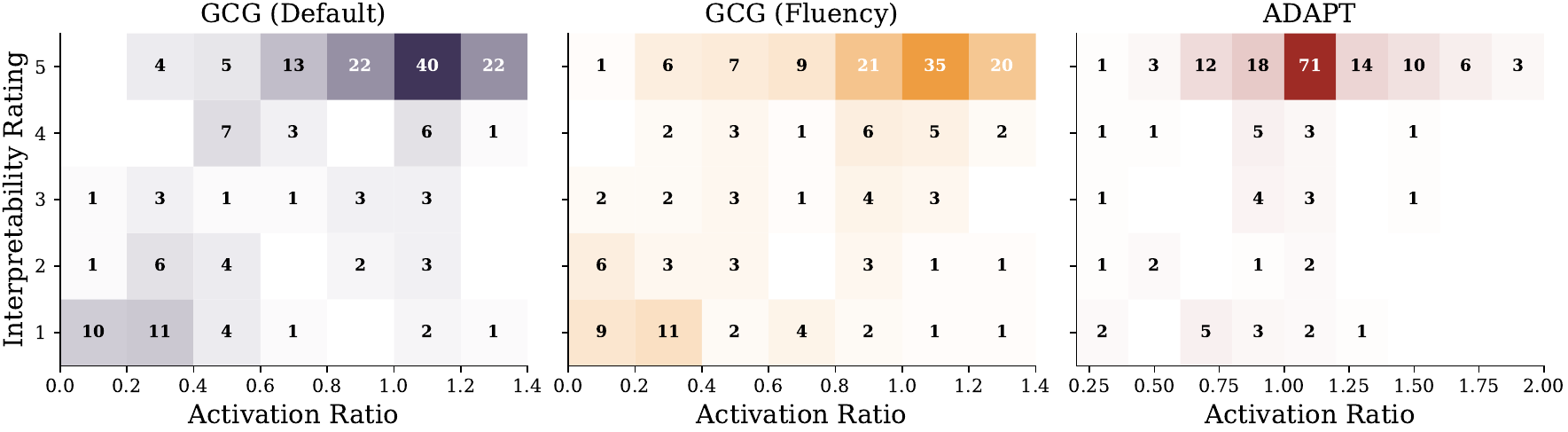}
    \caption{Rank/rating co-occurrence matrix for GCG, GCG with a $\lambda=1.0$ fluency penalty, and ADAPT, respectively.}
    \label{fig:rating_matrix}
\end{figure}

\begin{figure}
    \centering
    \includegraphics[width=1.0\linewidth]{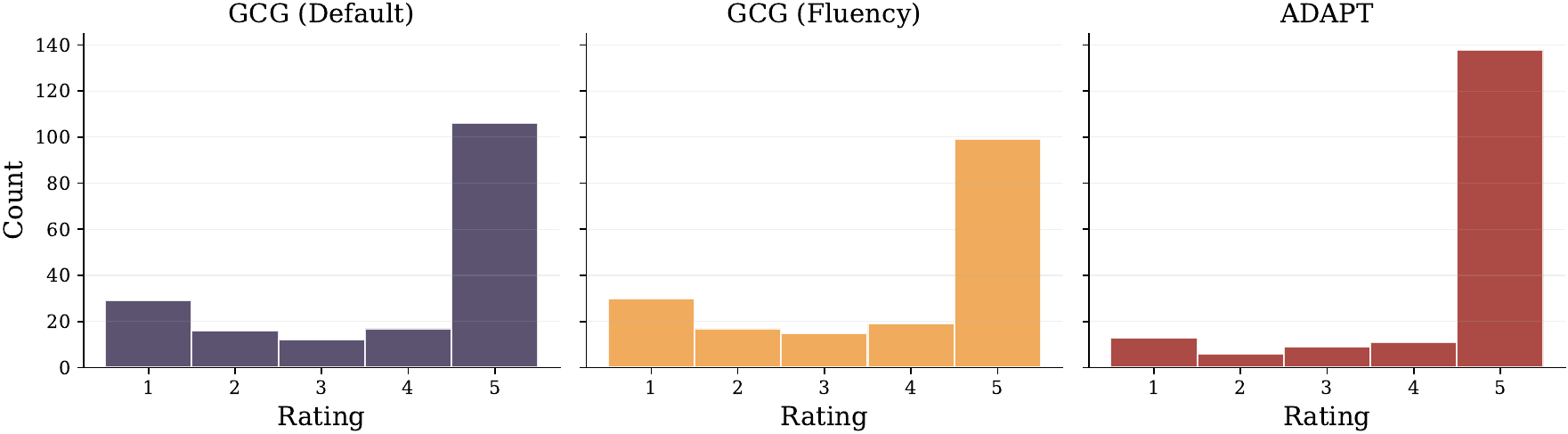}
    \caption{Rating distribution for GCG, GCG with a $\lambda=1.0$ fluency penalty, and ADAPT, respectively.}
    \label{fig:rating_dist}
\end{figure}

Figure \ref{fig:rating_matrix} shows that, for GCG, higher activation reliably correlates (Pearson $r=0.645, p<10^{-5}$) with more interpretable prompts, despite the lack of fluency constraints. Indeed, high activation (ratio$>1.0$) prompts seem overwhelmingly interpretable, whereas the correlation is less clear for lower activation regimes. The $\lambda = 1.0$ run shows similar patterns, with no remarkable improvement in interpretability. Indeed, as Figure \ref{fig:rating_dist} shows, the interpretability distribution is remarkably similar between the two methods, with the penalty appearing to slightly compress the rating distribution rather than shift it toward higher interpretability. 

\begin{figure}
    \centering
    \includegraphics[width=0.8\linewidth]{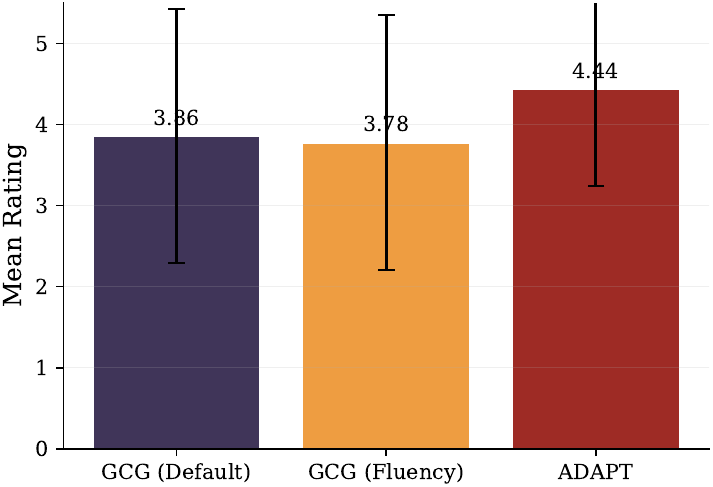}
    \caption{Quality rating for each method averaged over all latents and annotators.}
    \label{fig:method_rating}
\end{figure}

Our analysis also establishes that ADAPT is significantly better at generating interpretable prompts. Figure \ref{fig:method_rating} shows the mean rating for each of the methods, establishing the superiority of ADAPT (Wilcoxon Statistic of $4459.5$, $p < 10^{-8}$), while showing that GCG with $\lambda = 1.0$ does not produce significant changes in human-rated interpretability. Lastly, we note that because the activation ratio is a mere proxy for human interpretability, these results provide the most compelling evidence in support of our design decisions in ADAPT.

\section{Extensions}
\label{sec:extensions}

ADAPT naturally extends to objectives involving multiple latents by summing their activations. We conducted a brief experiment with dual latent optimization to illustrate this extension potential. Following \citet{shabalin_evolutionary_2024}, we optimize dual-latent objectives using an $L_{0.1}$ norm, which encourages balanced activation of both latents rather than maximizing one at the expense of the other:
\[
\mathcal{L}(t) = \left\| \begin{pmatrix} f_i(t) \\ f_j(t) \end{pmatrix} \right\|_{0.1},
\]
where $f_i(t)$ and $f_j(t)$ are the activation magnitudes of the two target latents for prompt $t$.

We found not all pairs of latents are amenable to optimization, since their associated patterns can be mutually exclusive. For instance, if both are high in locality (firing for narrow subsequences), it might not be possible to obtain high simultaneous activations on the same position. Notwithstanding, we achieved success in jointly optimizing pairs where one latent was local and the other fired for wide contexts. Table \ref{tab:multi} shows combinations of 3 different layer 25 residual stream latents, and their respective optimized prompts. Latent \href{https://www.neuronpedia.org/gemma-2-2b/25-gemmascope-res-16k/823}{823} is a low locality latent which fires for contexts related to the Marvel Cinematic Universe. Latent \href{https://www.neuronpedia.org/gemma-2-2b/25-gemmascope-res-16k/2738}{2738} is highly local and activates for terms like ``bit'' and ``little''. Latent \href{https://www.neuronpedia.org/gemma-2-2b/25-gemmascope-res-16k/1264}{1264} is local and fires for the token ``syn'' and spelling variants. Lastly, latent \href{https://www.neuronpedia.org/gemma-2-2b/25-gemmascope-res-16k/154}{154} is low locality and is active for text describing magnets and their properties. ADAPT is able to find interesting combinations of the patterns encoded by each latent which are interpretable, most notably the last example where the subsequence ``Stark magnet'' concisely fits the MCU (Tony Stark) and magnet patterns.

\begin{table}[h]
\centering
\small
\begin{tabular}{@{}ll p{4.2cm}@{}}
\toprule
Latent A & Latent B & Optimized Prompt \\
\midrule
823 (MCU) & 2738 (bit) & \texttt{nearby Avenger/avenulla Rd Avengers is bits} \\
1264 (syn) & 154 (magnets) & \texttt{ul magnetism currently ( determinediuminium, syn} \\
823 (MCU) & 154 (magnets) & \texttt{times aanki Stark magnet first unveiled as the} \\
\bottomrule
\end{tabular}
\caption{Optimized prompts for dual-latent objectives. All latents from layer 25. Prompts successfully activate both target latents, with local features (2738, 1264) appearing as specific tokens and low-locality features (823, 154) reflected in semantic context.}
\label{tab:multi}
\end{table}

The results presented here suggest that extending ADAPT to visualization of circuits or other complex objectives would prove fruitful, though we leave this to future work. 

\section{Conclusions}
\label{sec:conclusion}

We presented ADAPT, a hybrid prompt optimization method for LLM feature visualization. Through evaluation on SAE latents from Gemma 2 2B, we identified key failure modes of existing methods: GCG's gradient-based estimates exhibit poor signal-to-noise ratios and high sensitivity to initialization, while gradient-free beam search provides complementary exploration capabilities but lacks the fine-grained refinement needed for convergence.

ADAPT addresses these limitations through three principal design choices: (1) beam-search initialization to start optimization in favorable regions of prompt space, (2) adaptive mixing of gradient-guided and logit-based mutations to balance exploitation with exploration, and (3) a diversity-preserving slot management system that maintains independent optimization trajectories to avoid premature convergence to local minima.

Our experiments demonstrate that ADAPT outperforms GCG on 70.4\% of features ($p < 10^{-28}$) and achieves consistently strong performance across all model layers, unlike prior methods whose effectiveness varies with depth. Importantly, our manual evaluation reveals that high activation correlates strongly with human-rated interpretability ($r=0.645$), and that ADAPT produces significantly more interpretable prompts than baselines, providing direct evidence that optimizing for activation recovers meaningful feature patterns rather than exploiting artifacts.

These results establish that feature visualization for LLMs is tractable, but requires design assumptions tailored to the domain's unique challenges: discrete optimization over a vast token space, interference between non-orthogonal latent directions, and a loss landscape prone to local minima. We hope ADAPT provides a useful tool for interpretability research, and welcome extensions for visualization of steering vectors, circuit components, and other latent directions where exhaustive dataset search is prohibitive.

\paragraph{Limitations and Future Work.} Our evaluation focused on residual stream SAEs from a single model family. Extending ADAPT to larger models and more diverse latent extraction methods remains as possible future work. Additionally, while our metrics provide rigorous comparison, integration with automated interpretability pipelines that correlate with human judgment \cite{paulo2025evaluatingsaeinterpretabilityexplanations} would enable larger-scale evaluation.

\bibliographystyle{icml2025}
\bibliography{cleaned_bib}

\newpage
\appendix

\section{Appendix}
\label{sec:appendix}

\subsection{Implementation Details}
\label{sec:appendix_implementation}

Table \ref{tab:hyperparams} provides detailed hyperparameter settings for all methods evaluated in this work.

\begin{table}[h]
\centering
\small
\begin{tabular}{@{}lcccc@{}}
\toprule
\textbf{Parameter} & \textbf{GCG} & \textbf{BEAST} & \textbf{EPO} & \textbf{ADAPT} \\
\midrule
\multicolumn{5}{l}{\textit{General}} \\
Iterations & 50 & 10 & 50 & 25 \\
Prompt length & 10 & 10 & 8--15 & 10 \\
Parallel prompts & 6 & -- & 6 & 10 \\
\midrule
\multicolumn{5}{l}{\textit{Candidate Generation}} \\
Candidates/iter & 32 & -- & 32 & 32 \\
Top-$k$ (gradient) & 512 & -- & 512 & 512 \\
Beam size & -- & 128 & -- & 30 \\
Expansion factor & -- & 32 & -- & 32 \\
\midrule
\multicolumn{5}{l}{\textit{Operation Probabilities}} \\
GCG-swap & 1.0 & -- & 0.50 & 0.50 \\
Logit-swap & -- & -- & 0.20 & 0.50 \\
Insert & -- & -- & 0.15 & -- \\
Delete & -- & -- & 0.15 & -- \\
\midrule
\multicolumn{5}{l}{\textit{Fluency}} \\
$\lambda_{\min}$ & -- & -- & 0.1 & 0.0 \\
$\lambda_{\max}$ & -- & -- & 15.0 & 3.0 \\
\midrule
\multicolumn{5}{l}{\textit{ADAPT-specific}} \\
Diversity groups & -- & -- & -- & 3 \\
Guaranteed slots/group & -- & -- & -- & 2 \\
Init sequences & -- & -- & -- & 5000 \\
\bottomrule
\end{tabular}
\caption{Hyperparameter settings for all evaluated methods.}
\label{tab:hyperparams}
\end{table}

Algorithm \ref{alg:adapt} presents pseudocode for ADAPT.

\begin{algorithm}[h]
\caption{ADAPT optimization procedure}
\label{alg:adapt}
\begin{algorithmic}[1]
\REQUIRE Target latent $f$, model $M$, SAE $S$, iterations $T$, prompts $K$, groups $G$
\ENSURE Optimized prompt $t^*$
\STATE $\mathcal{P} \gets \textsc{BeamSearchInit}(f, M, S, G)$ \COMMENT{$K$ prompts across $G$ groups}
\FOR{$i = 1$ \TO $T$}
    \STATE Sample operation $\text{op} \sim \text{Bernoulli}(p_{\text{gcg}})$
    \IF{$\text{op} = \textsc{GCG-Swap}$}
        \STATE Compute gradient estimate $\nabla$ via one-hot embedding
        \STATE $\mathcal{C} \gets \textsc{GradientMutate}(\mathcal{P}, \nabla)$ \COMMENT{Low-gradient tokens}
    \ELSE
        \STATE $\mathcal{C} \gets \textsc{LogitMutate}(\mathcal{P})$ \COMMENT{Model's predictions}
    \ENDIF
    \FOR{each candidate $c \in \mathcal{C}$}
        \STATE $a_c \gets f(c)$; \quad $\lambda_i \gets \lambda_{\max} / (1 + e^{-s(i - t_{\text{mid}})})$
        \STATE $\text{score}_c \gets a_c - \lambda_i \cdot \max(0, \text{CE}(c) - \tau)$
    \ENDFOR
    \FOR{$g = 1$ \TO $G$}
        \STATE Keep top candidates from group $g$ in guaranteed slots
    \ENDFOR
    \STATE Fill remaining slots globally by score (with per-group cap)
    \STATE $\mathcal{P} \gets$ selected candidates
\ENDFOR
\RETURN $\arg\max_{t \in \mathcal{P}} f(t)$
\end{algorithmic}
\end{algorithm}

\subsection{Latent Selection Procedure}
\label{sec:appendix_latent_selection}

We computed proxies for the three taxonomic axes described by Graham et al. using data available via the Neuronpedia API.

\paragraph{Density.} A latent's density corresponds to the frequency with which it activates. We take the fraction of tokens for which the latent is active, which is directly provided by the Neuronpedia API as \texttt{frac\_nonzero}.

\paragraph{Locality.} Locality describes how concentrated a latent's activation is within a sequence. A highly local latent exhibits high activation for a single token with negligible activation for surrounding tokens, whereas a low-locality latent may activate across entire phrases. Given a set of $N$ maximally activating examples for a latent, let $a^{(j)} = (a^{(j)}_1, \ldots, a^{(j)}_{L_j})$ denote the activation values across all $L_j$ tokens in example $j$. We define locality as:
$$
\text{Locality} = \frac{1}{N} \sum_{j=1}^{N} \frac{\max_i a^{(j)}_i}{\sum_i a^{(j)}_i}
$$
For a highly local latent, this ratio approaches 1, as the maximally activating token captures most of the activation mass.

\paragraph{Diversity.} Diversity indicates the variety of tokens a latent responds to. A low-diversity latent fires for a specific token, while a high-diversity latent activates for many different tokens sharing a semantic or functional role. Given the top $N$ maximally activating examples, let $T = \{t_1, \ldots, t_N\}$ be the set of tokens achieving maximal activation in each example. We define diversity as:
$$
\text{Diversity} = \frac{|\text{lowercase}(T)|}{N}
$$
where $\text{lowercase}(T)$ applies lowercasing to all tokens before computing the unique set size. This normalization accounts for capitalization variants of the same underlying token.

\subsection{Human Evaluation Protocol}
\label{sec:appendix_human_eval}

\begin{figure*}[p]
    \centering
    \includegraphics[width=\textwidth]{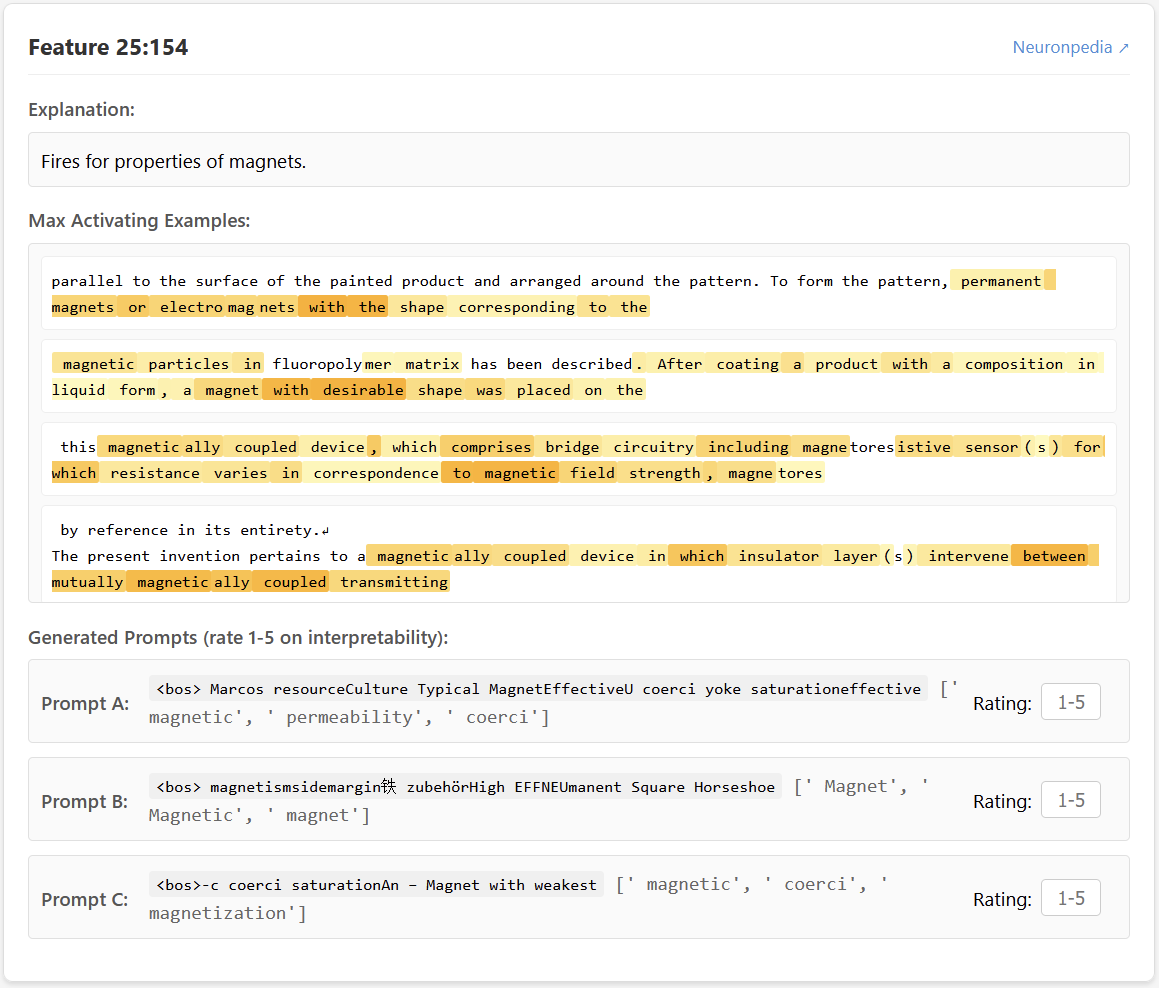}
    \caption{Annotation task example for the `magnet' latent.}
    \label{fig:annotation_example}
\end{figure*}

Annotators were presented with the following information for each latent: (1) an explanation of what the latent detects, (2) maximally activating dataset examples with highlighted tokens showing activation strength, and (3) three generated prompts from different methods in randomized order, with the 3 most likely token completions in square brackets to aid in rating `Say X' features. Annotators could also access interactive exploration via Neuronpedia. Figure \ref{fig:annotation_example} shows a single latent's annotation interface.

Prompts were rated on the following 1--5 scale:
\begin{enumerate}
    \item \textbf{No discernible pattern}: The prompt appears random or unrelated to the latent's function.
    \item \textbf{Semantic ballpark}: The prompt is in the general semantic area of the reference examples but is otherwise uninterpretable.
    \item \textbf{Pattern understandable}: The pattern the prompt is targeting is understandable given context, but not discernible from the optimized prompt alone.
    \item \textbf{Pattern present but different}: The pattern is clearly present but differs in some way from reference examples (e.g., different language, domain, or format).
    \item \textbf{Pattern obvious and matches}: The pattern is immediately obvious and closely matches the maximally activating dataset examples.
\end{enumerate}

\end{document}